\documentclass[twoside]{article}

\usepackage{PRIMEarxiv}

\usepackage[utf8]{inputenc} 
\usepackage[T1]{fontenc}    
\usepackage{hyperref}       
\usepackage{url}            
\usepackage{booktabs}       
\usepackage{amsfonts}       
\usepackage{nicefrac}       
\usepackage{microtype}      
\usepackage{lipsum}
\usepackage{fancyhdr}       

\usepackage{amsmath}
\usepackage[colorinlistoftodos]{todonotes}
    
\usepackage{subcaption}

\usepackage{amssymb} 
\usepackage{tcolorbox} 
\usepackage{makecell}
\PassOptionsToPackage{ruled,vlined,linesnumbered}{algorithm2e}
\usepackage{algorithm2e}
\usepackage{booktabs} 
\usepackage{microtype}
\usepackage{graphicx}
\usepackage{booktabs} 
\usepackage{wrapfig}
\usepackage{algorithmic}

\title{Visual Reasoning Agent: Robust Vision Systems in Remote Sensing via Inference-Time Scaling
}

\author{Chung-En Johnny Yu \\
    University of West Florida \\
    Pensacola, Florida, USA \\
    \texttt{cy31@students.uwf.edu} \\
    \And
    Brian Jalaian \\
    University of West Florida \\
    Pensacola, Florida, USA  \\
    \texttt{bjalaian@uwf.edu} 
    \And
    Nathaniel D. Bastian \\
    United States Military Academy \\
    West Point, New York, USA  \\
    \texttt{nathaniel.bastian@westpoint.edu} 
}

\begin{document}
\maketitle

\begin{abstract}
Building robust vision systems for high-stakes domains such as remote sensing requires stronger visual reasoning than what single-pass inference typically provides; yet, retraining large models is often computationally expensive and data intensive.
We present \textbf{Visual Reasoning Agent (VRA)}, a training-free agentic visual reasoning framework that orchestrates off-the-shelf large vision-language models (LVLMs) with a large reasoning model (LRM) through an iterative \emph{Think-Critique-Act} loop for cross-model verification, self-critique, and recursive refinement. 
On the remote sensing benchmark VRSBench VQA dataset, VRA consistently outperforms multiple standalone LVLM baselines and achieves up to 40.67\% improvement on challenging question types spanning both perception and reasoning tasks.
In addition, integrating three LVLMs with VRA improves the overall accuracy of the standalone LVLMs from 52.8\% to 78.8\%, demonstrating the effectiveness of agentic reasoning with increased inference-time compute.
\end{abstract}


\section{Introduction}

\begin{wrapfigure}{r}{0.5\columnwidth}
    \vspace{-15pt}
    \centering
    \includegraphics[width=0.5\textwidth]{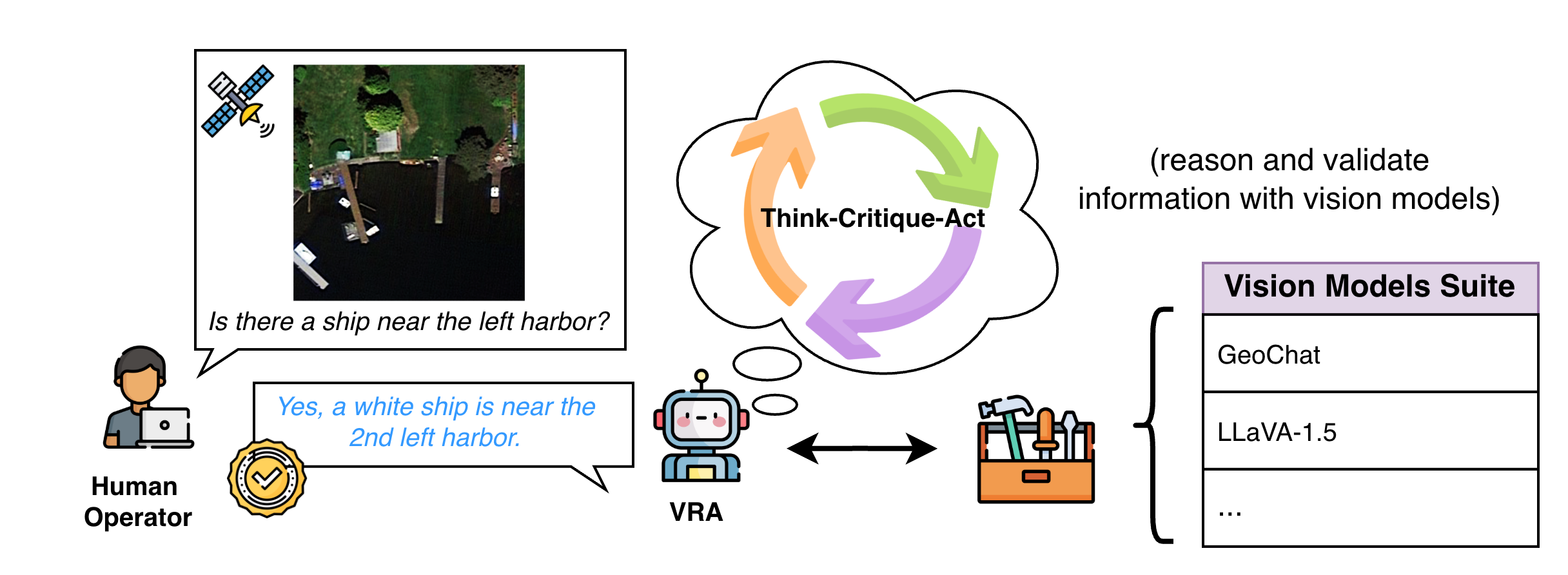}
    \caption{Overview of the Visual Reasoning Agent (VRA) framework. VRA performs iterative \emph{Think-Critique-Act} reasoning with cross-model verification to improve reasoning on remote sensing tasks.}
    \label{fig:vra_overview}
    \vspace{-10pt}
\end{wrapfigure}

Artificial intelligence (AI)-enabled vision systems, including pure vision models like object detectors, and large vision-language models (LVLMs), often struggle to generalize reliably in high-stakes domains such as remote sensing, where tasks frequently require both fine-grained perception and multi-step visual reasoning \cite{liang2024survey}.
While fine-tuning can improve performance, it is often computationally expensive, requires extensive labeled data, and offers no guarantee of improved robustness. 
Consequently, many practitioners lack the resources or domain-specific datasets needed to retrain models for operational deployment.

Recent research has explored improving LVLM factuality through post-hoc correction methods such as Visual Fact Checker \cite{ge2024visual}, LogicCheckGPT \cite{wu2024logical}.
However, these approaches typically rely on rigid, predefined pipelines that are difficult to adapt to diverse visual reasoning tasks. 
More broadly, current LVLM systems frequently generate single-pass outputs, which can lead to brittle reasoning and hallucinated interpretations when confronted with complex visual scenes.
Agentic reasoning frameworks offer a promising alternative. 
Recent advances in large language model (LLM)-based agentic reasoning, e.g., ReAct \cite{yao2022react} and CRITIC \cite{gou2023critic}, demonstrate that iterative reasoning and self-critique can significantly improve robustness and reasoning quality.
These approaches allow systems to refine model outputs through multi-step reasoning and tool interaction. 
However, while agentic methods have been widely studied in text-based tasks, their application to improving robustness across diverse LVLM visual reasoning tasks remains relatively underexplored \cite{chen2025towards}.

To address these challenges, we introduce the \textbf{Visual Reasoning Agent (VRA)}, an agentic framework that enhances LVLM reasoning through an iterative \emph{Think-Critique-Act} loop. 
As illustrated in Figure.~\ref{fig:vra_overview}, VRA orchestrates multiple LVLMs to provide diverse visual evidence while a large reasoning model (LRM) performs structured reasoning, self-critique, and follow-up querying. 
This design enables cross-model verification and iterative refinement of visual understanding, improving robustness without requiring model retraining.

\begin{figure*}[t]
    \centering
    \includegraphics[width=0.9\linewidth]{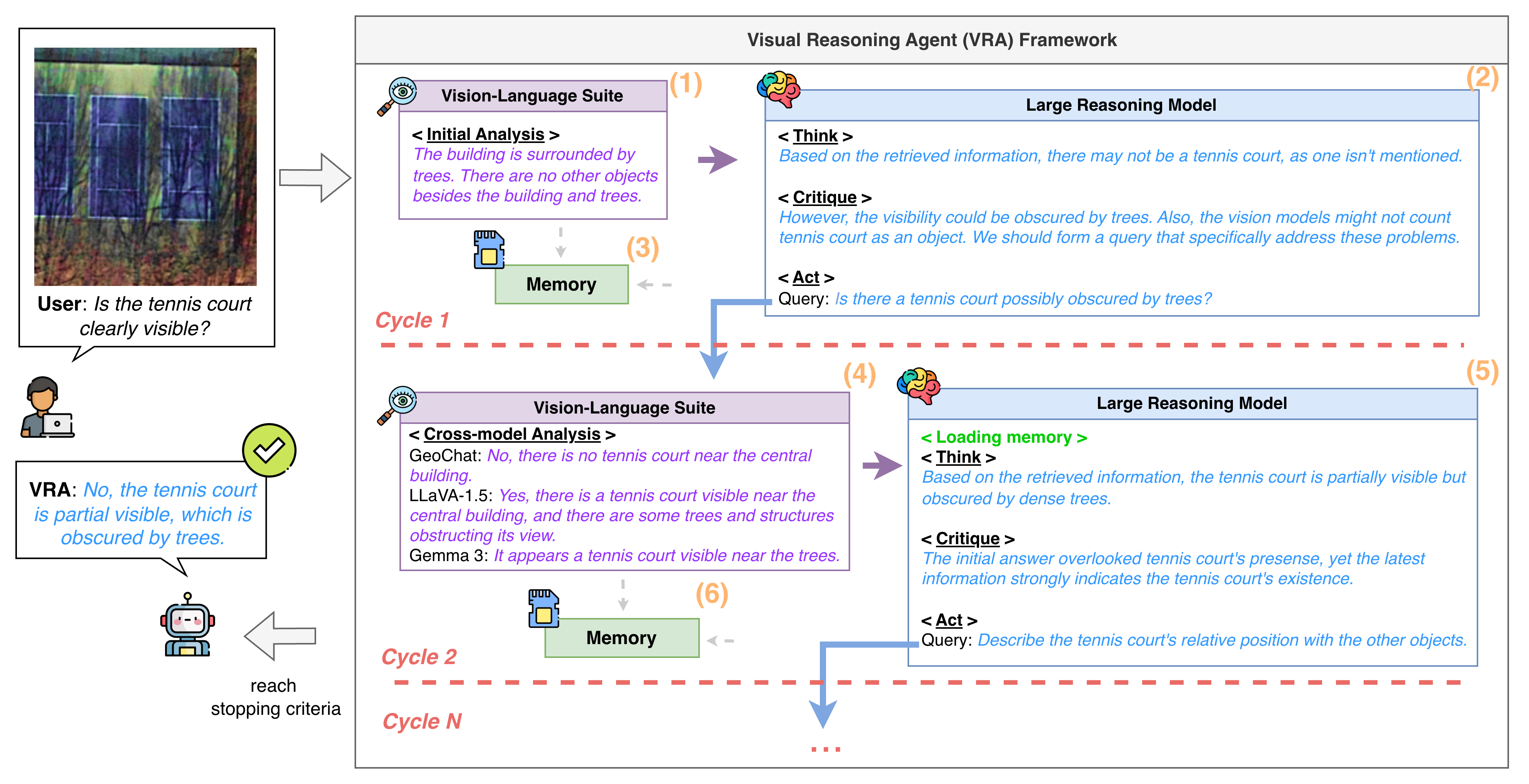}
    \caption{Workflow of Visual Reasoning Agent (VRA) framework. VRA performs iterative visual reasoning via a Think-Critique-Act loop, where heterogeneous LVLMs provide diverse visual evidence, a Large Reasoning Model (LRM), e.g., QwQ, critiques and refines reasoning and generates follow-up queries, and a memory module stores visual evidence and reasoning trajectory across iterations.}
    \label{fig:vra_workflow}
\end{figure*}

Remote sensing presents particularly demanding conditions for visual reasoning, including high-resolution imagery, dense small objects, long-range spatial dependencies, and frequent domain shifts across sensors and geographic regions.
Recent LVLMs for remote sensing, such as RSGPT \cite{wu2023rsgpt} and GeoChat \cite{kuckreja2024geochat}, demonstrate promising capabilities but still rely primarily on single-pass inference. 
Benchmarks such as VRSBench \cite{li2024vrsbench}, XLRS-Bench \cite{liu2025xlrsbench}, and SkySense-O \cite{liu2025skysenseo} further highlight persistent challenges in fine-grained perception and multi-step reasoning for satellite imagery understanding. 
Our proposed framework addresses these limitations by enabling iterative reasoning with cross-model validation, producing more reliable outputs than conventional single-pass LVLM inference. 
In addition, VRA records reasoning traces and visual evidence throughout the reasoning process, providing transparent and auditable decision flows for reliable AI deployment.

Our key contributions are summarized as follows:
\begin{enumerate}
    \item We introduce \textbf{Visual Reasoning Agent (VRA)}, an iterative \emph{Think-Critique-Act} framework that coordinates multiple LVLMs and a reasoning model to improve visual reasoning robustness.
    \item VRA integrates heterogeneous LVLMs to provide complementary visual evidence, enabling cross-model validation that reduces hallucination and model-specific bias.
    \item Experiments on the VRSBench VQA benchmark demonstrate up to 40\% accuracy improvements across challenging remote sensing reasoning tasks.
\end{enumerate}


\section{Visual Reasoning Agent}
\label{sec:vra_method}

We propose the Visual Reasoning Agent (VRA) framework to improve the robustness of large vision-language models (LVLMs) across diverse visual tasks, encompassing both perception and reasoning-intensive visual tasks.
VRA adopts an iterative \emph{Think-Critique-Act} loop, inspired by ReAct \cite{yao2022react} and Reflexion \cite{shinn2023reflexion}, enabling structured self-refinement and cross-model validation at inference time. 
As illustrated in Figure~\ref{fig:vra_workflow}, the framework consists of three core components: (i) a Vision-Language Suite, (ii) a Large Reasoning Model (LRM), and (iii) a memory module.

\subsection{Framework Components}

\textbf{Vision-Language Suite.}
The Vision-Language Suite comprises a set of heterogeneous LVLMs that independently analyze the input image and provide diverse natural-language descriptions. 
By aggregating multiple model perspectives, the system mitigates reliance on any single LVLM and reduces model-specific bias or hallucination.
\textbf{Large Reasoning Model (LRM).}
The LRM functions as the central agent responsible for reasoning over the collective visual evidence. 
It is selected for strong agentic reasoning capability and structured multi-step reasoning outputs. 
Beyond answer generation, the LRM performs self-critique to detect inconsistencies across model responses and contradictions within its own reasoning trajectory.
For example, we utilized QwQ \cite{qwq32b} in our experiments.
\textbf{Memory Module.}
The memory module logs the visual observations and reasoning traces across iterations. 
This module serves three purposes:
(1) maintaining logical consistency across cycles,
(2) preventing redundant queries or reasoning steps, and
(3) providing auditable reasoning traces for post-hoc verification by human operators.

\subsection{VRA Workflow}

As illustrated in Figure~\ref{fig:vra_workflow}, the workflow proceeds iteratively:
\textbf{(1) Initial Visual Context.}
An LVLM generates an initial image analysis, providing foundational visual context and scene interpretation.
\textbf{(2) Think-Critique-Act (Cycle $1$).}
The LRM performs three structured operations:
    \emph{Think}: Reasons out an initial answer based on the current visual information.
	\emph{Critique}: Identifies potential inconsistencies in visual information and contradictions in its reasoning.
	\emph{Act}: Generates a targeted follow-up query to retrieve additional visual evidence from the Vision-Language Suite.
\textbf{(3) Memory Update.}
The visual information generated by the LVLM and the Think-Critique-Act outputs by the LRM are stored in the memory module. 
The memory is used for the LRM reasoning process in Think-Critique-Act loop in the subsequent cycles.
\textbf{(4) Cross-Model Analysis (Cycle $\geq 2$).}
The heterogeneous LVLMs respond to the LRM-generated query. 
This step performs cross-model validation, capturing diverse interpretations of the same visual evidence.
\textbf{(5) Iterative Think-Critique-Act with Memory Loading.}
The LRM retrieves the entire reasoning trajectory from memory module and performs another Think-Critique-Act cycle based on the updated visual evidence. 
By conditioning on the complete reasoning trajectory, the agent enforces consistency and refines its answer across iterations.
\textbf{(6) Memory Update and Continuation.}
The latest visual information and reasoning trace are stored in the memory module. 
The iterative process continues until a stopping criterion is met, such as reaching a maximum number of iterations.

VRA is motivated by the hypothesis that increasing inference-time compute through structured agentic reasoning improves robustness for vision systems. 
Rather than relying on single-pass inference from one LVLM, VRA performs: cross-model verification, structured self-critique, iterative evidence refinement.
This design is particularly beneficial for complex visual reasoning tasks, where superficial single-model inference is insufficient. 
Although computational cost increases, the trade-off is justified in high-stakes applications such as remote sensing, surveillance analysis, and safety-critical decision support, where reliability and robustness outweigh throughput.

The detailed procedure of VRA is summarized in Algorithm~\ref{alg:vra}.
The detailed prompt templates used in VRA are provided in Appendix~\textit{Prompt Templates}.

\begin{algorithm}[h]
\caption{\textbf{Visual Reasoning Agent (VRA)}}
\label{alg:vra}
\small
\begin{algorithmic}[1]
\REQUIRE Image $I$, question $Q$, Vision-Language Suite $\mathcal{V}=\{V_1,\dots,V_M\}$, LRM $R$, memory $\mathcal{M}$, max iterations $K$
\ENSURE Final answer $\hat{a}$ and trace $\mathcal{M}$

\STATE $\mathcal{M}\leftarrow \emptyset$
\STATE \textbf{Initial visual context:} $o^{(0)} \leftarrow V_1(I)$
\STATE $\mathcal{M}\leftarrow \mathcal{M}\cup\{o^{(0)}\}$

\FOR{$t=1$ \TO $K$}
    \STATE \textbf{Think-Critique-Act:} $(\hat{a}^{(t)}, c^{(t)}, q^{(t)}) \leftarrow R(Q, \mathcal{M})$ \\
    \qquad $\triangleright$ Reason out answer, critique, and follow-up query
    \STATE $\mathcal{M}\leftarrow \mathcal{M}\cup\{\hat{a}^{(t)}, c^{(t)}, q^{(t)}\}$
    \STATE \textbf{Cross-model visual evidence:} $O^{(t)} \leftarrow \{V_m(I, q^{(t)})\}_{m=1}^{M}$
    \STATE $\mathcal{M}\leftarrow \mathcal{M}\cup\{O^{(t)}\}$
\ENDFOR

\STATE $\hat{a}\leftarrow \hat{a}^{(t)}$
\RETURN $\hat{a}, \mathcal{M}$
\end{algorithmic}
\end{algorithm}
\section{Preliminary Experiments and Results}


\subsection{Experiment Setup}
\label{sec:exp_setup}

To validate the effectiveness of VRA, we conducted preliminary experiments on visual reasoning tasks in the remote sensing domain, a high-stakes application area where robustness is critical for applications like disaster response and surveillance. 
\textbf{Dataset.} 
We evaluated VRA on the VRSBench VQA dataset \cite{li2024vrsbench}, which offers comprehensive satellite imagery understanding tasks that evaluate the capabilities of both fine-grained perception and complex visual reasoning. 
The question types in VRSBench are meticulously categorized to cover both low-level visual attributes—such as object category, count, color, and shape—and high-level visual reasoning, including spatial relationships and scene-level understanding. 
This hierarchical structure allows for a robust assessment of a model's ability to transition from basic feature identification to complex relational logic in a geo-spatial context.
We sampled 50 questions for each of the ten question types, resulting in 500 questions in total.
Performance is measured by accuracy per question type and overall accuracy is computed as the average across all evaluated question types.
\textbf{Models.} 
To demonstrate VRA's adaptability and robustness, we integrated it with multiple established LVLMs as plug-in components. 
These baseline LVLMs include: GeoChat (7B) \cite{kuckreja2024geochat}, LLaVA-v1.5 (7B) \cite{liu2024improved}, and Gemma 3 (12B) \cite{team2025gemma}. 
Note that GeoChat is a specialized RS-LVLM, fine-tuned LLaVA-v1.5 on remote sensing datasets. 
The LRM is QwQ (32B) \cite{qwq32b}, a text-based reasoning model that enables agentic operations. 
Also, Phi-4 \cite{abdin2025phi} is used in VRSBench VQA dataset for evaluating free-form answers.
The prompt template used for evaluation is provided in Appendix~\textit{Prompt Templates}.
\textbf{VRA Variants.}
VRA is designed to be modular, allowing for the integration of single or multiple off-the-shelf LVLMs as visual information providers. 
We define the following VRA variants for comparison: 
    \textit{VRA (LVLM\_NAME)}, which denotes VRA with only the specified LVLM as the visual information source, e.g., GeoChat, LLaVA-1.5, Gemma 3. 
    \textit{VRA (+1 LVLM avg)} represents the average of VRA (GeoChat), VRA (LLaVA-1.5), and VRA (Gemma 3), which is used in Figure \ref{fig:qtype_acc} and \ref{fig:runtime}. 
    \textit{VRA (+2 LVLMs)} utilizes GeoChat and LLaVA-1.5 in VRA. 
    \textit{VRA (+3 LVLMs)} integrates all three baseline LVLMs (GeoChat, LLaVA-1.5, and Gemma 3) concurrently.
    Finally, \textit{LVLM alone (avg)} in the figures refers to the average performance of the three baseline LVLMs (GeoChat, LLaVA-1.5, and Gemma 3) without VRA. 


\subsection{Results Analysis}

\begin{wrapfigure}{r}{0.5\columnwidth}
    \vspace{-10pt}
    \centering
    \includegraphics[width=0.95\linewidth]{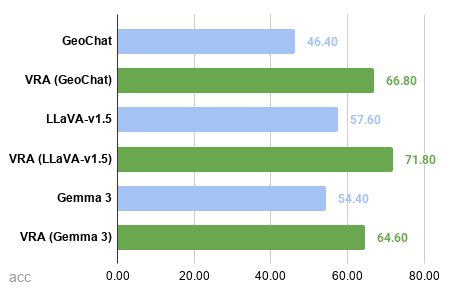}
    \caption{Overall accuracy comparison between standalone LVLMs and VRA-augmented variants.}
    \label{fig:acc}
    \vspace{-10pt}
\end{wrapfigure}

\textbf{Overall Accuracy Improvement.}
As illustrated in Figure~\ref{fig:acc}, VRA consistently improves the overall accuracy across all evaluated LVLMs. GeoChat improves from 46.40\% to 66.80\% (+20.40\%), LLaVA-1.5 increases from 57.60\% to 71.80\% (+14.20\%), and Gemma 3 rises from 54.40\% to 64.60\% (+10.20\%). These consistent gains demonstrate that the proposed agentic reasoning framework effectively enhances the reliability of LVLM predictions. By iteratively refining reasoning through the Think-Critique-Act loop and incorporating cross-model visual evidence, VRA mitigates errors produced by single-pass LVLM inference.

\begin{figure}[t]
    \centering
    \includegraphics[width=0.5\linewidth]{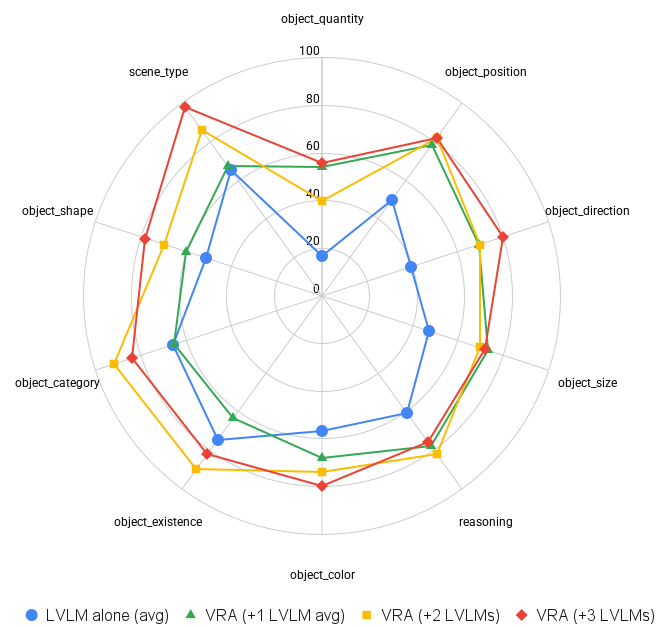}
    \caption{Performance of averaged standalone LVLMs versus VRA-augmented variants across different question types. Refer to Section~\ref{sec:exp_setup} for the definitions of VRA variants.}
    \label{fig:qtype_acc}
\end{figure}

\textbf{Enhanced Performance Across Question Types.}
Figure~\ref{fig:qtype_acc} and Table~\ref{tab:vrsbench_accuracy_new} further analyze performance across ten visual reasoning categories. 
VRA consistently outperforms the baseline LVLM average across most question types, particularly in tasks requiring precise spatial understanding and reasoning. 
The most significant improvement is observed in \textit{object direction} question type, where accuracy increases from 39.33\% to 80.00\% (+40.67\%) when using three LVLMs. 
Substantial gains are also observed in \textit{object quantity}, where accuracy increases from 16.67\% to 56.00\% (+39.33\%) and \textit{scene type}, where increases from 65.33\% to 98.00\% (+32.67\%). 
These improvements suggest that cross-model verification and iterative reasoning are particularly beneficial for tasks requiring spatial interpretation and quantity understanding. 
Performance improvements are also consistent in reasoning-heavy categories such as \textit{object position}, \textit{object size}, and \textit{reasoning}, indicating that VRA enhances both perception-level and reasoning-level visual question answering capabilities. 
Overall accuracy increases from 52.80\% for standalone LVLMs to 78.80\% (+26\%) when VRA integrates three LVLMs.

\textbf{Impact of Increasing Multiple LVLMs.}
Increasing the number of LVLMs in VRA further improves performance. 
As shown in Table~\ref{tab:vrsbench_accuracy_new}, overall accuracy increases from 67.73\% (+1 LVLM) to 75.60\% (+2 LVLMs) and 78.80\% (+3 LVLMs) with three LVLMs. 
This trend highlights the benefit of cross-model analysis, where diverse visual interpretations provide complementary information that reduces individual model biases and hallucinations.

\textbf{Runtime Analysis}

\begin{wrapfigure}{r}{0.5\columnwidth}
    \vspace{-15pt}
    \centering
    \includegraphics[width=0.95\linewidth]{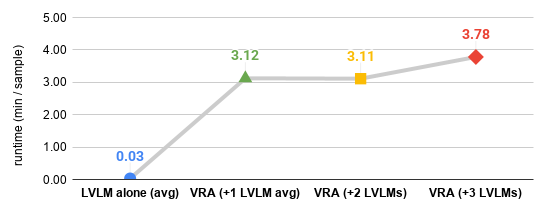}
    \caption{Overall runtime (minute/sample) comparison between averaged standalone LVLMs and VRA-augmented variants. Refer to Section~\ref{sec:exp_setup} for VRA variants.}
    \label{fig:runtime}
    \vspace{-10pt}
\end{wrapfigure}

Figure~\ref{fig:runtime} presents the runtime cost of the proposed framework. 
While standalone LVLM inference requires only 0.03 minutes per sample on average, incorporating the agentic reasoning loop increases the runtime to approximately 3.12 minutes with one LVLM and 3.78 minutes with three LVLMs. 
The increase arises from iterative reasoning cycles and multiple LVLM queries required for cross-model validation. 
Although VRA introduces higher inference-time cost, the trade-off is justified in applications where robustness and reliability are critical, such as remote sensing and safety-critical decision support.

\begin{table*}[t]
    \centering
    \scriptsize
    \setlength{\tabcolsep}{3pt} 
    \caption{Performance comparison between average standalone LVLMs and VRA variants. The metric is accuracy (\%)}
    \begin{tabular}{lccccccccccc}
        \toprule
        Model & obj\_quantity & obj\_position & obj\_direction & obj\_size & reasoning & obj\_color & obj\_existence & obj\_category & obj\_shape & scene\_type & \textbf{overall} \\
        \midrule
        LVLM alone (avg) & 16.67 & 50.00 & 39.33 & 47.33 & 60.67 & 56.67 & 74.67 & 66.00 & 51.33 & 65.33 & \textbf{52.80} \\
        VRA (+1 LVLM avg) & 54.00 & 78.67 & 69.33 & 73.33 & 78.00 & 68.00 & 63.33 & 65.33 & 60.00 & 67.33 & \textbf{67.73} \\
        VRA (+2 LVLMs) & 40.00 & 82.00 & 70.00 & 70.00 & 82.00 & 74.00 & 90.00 & 92.00 & 70.00 & 86.00 & \textbf{75.60} \\
        VRA (+3 LVLMs) & 56.00 & 82.00 & 80.00 & 72.00 & 76.00 & 80.00 & 82.00 & 84.00 & 78.00 & 98.00 & \textbf{78.80} \\
        \bottomrule
    \end{tabular}
    \label{tab:vrsbench_accuracy_new}
\end{table*}

\section{Discussion}

Our preliminary findings from VRA provide compelling evidence that agentic reasoning significantly enhances the general robustness of LVLMs across diverse visual tasks. 
The substantial accuracy improvements in remote sensing for both perceptual and reasoning tasks, coupled with VRA's architectural components, underscore the potential for more reliable LVLM deployment in safety-critical domains.


\subsection{Iterative Agentic Reasoning and Diverse Models}

Unlike static inference, the \emph{Think-Critique-Act} loop allows VRA to recursively refine its understanding and outputs, which enables it to navigate complex visual information and resolve inconsistencies. 
This dynamic adaptation is crucial for high accuracy across visual reasoning challenges, as evidenced by VRA's consistent outperformance of standalone LVLMs. The robustness gained is from iterative self-correction.

By leveraging multiple heteorgeneous LVLMs, VRA inherently mitigates risks associated with reliance on a single model. 
Each LVLM in VRA offers a distinct visual encoder and training bias, enabling cross-model validation by the LRM. 
This enhances factual consistency and reducing susceptibility to individual LVLM failures. 
Moreover, the LRM plays a pivotal role by contributing its extensive reasoning capabilities. 
VRA leverages its powerful ``thinking" to synthesize information from disparate sources, critiques, and formulate necessary follow-up questions. 
This enables effective visual reasoning even with basic initial prompts, highlighting that robustness stems from the agentic workflow, the underlying reasoning model, and effective orchestration of diverse vision models, also contributing to make the agent's decision-making process more transparent and auditable.





\subsection{Limitations and Future Directions}

Our VRA framework is not without limitations, primarily related to computational overhead. 
As shown in the runtime analysis in Figure~\ref{fig:runtime}, leveraging a long-thinking process from a LRM and increasing the number of integrated LVLMs leads to a substantial increase in inference time, which could be a barrier for real-time applications.

Future work will focus on:
(1) Investigating more efficient agentic frameworks that retain robustness while significantly reducing computational overhead, e.g., dynamic agent routing techniques or early stopping criterion when VRA reaches high confidence.
(2) Evaluating VRA on hallucination benchmarks, e.g., Throne \cite{kaul2024throne} for intrinsic errors assurance.
(3) Rigorously validating VRA's adversarial robustness against a wider array of state-of-the-art adversarial attacks, exploring agentic defense mechanisms.
(4) Further evaluating VRA's visual capabilities on diverse safety-critical domains to assess VRA's generalizability for real-world reliability.


\section{Conclusion}
Robust vision systems are essential for high-stakes domains such as remote sensing. 
We have introduced \textbf{Visual Reasoning Agent (VRA)}, a training-free agentic reasoning framework that wraps off-the-shelf LVLMs in a \emph{Think-Critique-Act} loop. 
VRA's training-free design democratizes access to robust vision deployments, even for teams lacking fine-tuning resources, such as labeled data and exhaustive compute. 
Moreover, its cross-model validation and iterative self-correction principles point toward promising robust visual reasoning capabilities. 
Empirical results on the remote sensing benchmark VRSBench VQA dataset demonstrate up to 40\% accuracy improvements, confirming that increased inference-time compute via agentic reasoning is a justifiable trade-off for reliability.
Future work will focus on dynamic agent routing and confidence-based early stopping to reduce inference overhead.

\section*{Acknowledgment}
This work was supported in part by the U.S. Military Academy (USMA) under Cooperative Agreement No. W911NF-23-2-0108. The views and conclusions expressed in this paper are those of the authors and do not reflect the official policy or position of the U.S. Military Academy, U.S. Army, U.S. Department of War, or U.S. Government.

\bibliographystyle{unsrt}  
\bibliography{reference}

\newpage
\onecolumn


\section{Appendix}
\subsection{Comprehensive Results Details}
\label{sec:vra_result_table}

The following tables present the comprehensive evaluation results on the VRSBench VQA subset, consisting of 50 questions per type and overall performance of all question types. Table \ref{tab:vrsbench_accuracy} reports the accuracy results (\%) across different question types, comparing various baseline LVLMs and their corresponding VRA-augmented variants. Table \ref{tab:vrsbench_runtime} summarizes the runtime statistics, illustrating the computational costs associated with each model configuration. These results provide a holistic view of the performance trade-offs between accuracy and efficiency across diverse vision tasks.

\begin{table*}[h]
    \centering
    \scriptsize
    \setlength{\tabcolsep}{3pt} 
    \caption{VRSBench VQA Subset (50 questions per type) Accuracy in \%}
    \begin{tabular}{lccccccccccc}
        \toprule
        Model & obj\_quantity & obj\_position & obj\_direction & obj\_size & reasoning & obj\_color & obj\_existence & obj\_category & obj\_shape & scene\_type & \textbf{overall} \\
        \midrule
        GeoChat & 10.00 & 68.00 & 34.00 & 22.00 & 60.00 & 24.00 & 84.00 & 76.00 & 34.00 & 52.00 & \textbf{46.40} \\
        VRA (GeoChat) & 52.00 & 76.00 & 70.00 & 74.00 & 86.00 & 50.00 & 78.00 & 70.00 & 62.00 & 50.00 & \textbf{66.80} \\
        LLaVA-1.5 & 30.00 & 42.00 & 36.00 & 60.00 & 56.00 & 68.00 & 78.00 & 74.00 & 60.00 & 72.00 & \textbf{57.60} \\
        VRA (LLaVA-1.5) & 48.00 & 88.00 & 72.00 & 72.00 & 72.00 & 80.00 & 70.00 & 74.00 & 62.00 & 80.00 & \textbf{71.80} \\
        Gemma 3 & 10.00 & 40.00 & 48.00 & 60.00 & 66.00 & 78.00 & 62.00 & 48.00 & 60.00 & 72.00 & \textbf{54.40} \\
        VRA (Gemma 3) & 62.00 & 72.00 & 66.00 & 74.00 & 76.00 & 74.00 & 42.00 & 52.00 & 56.00 & 72.00 & \textbf{64.60} \\
        VRA (+2 LVLMs) & 40.00 & 82.00 & 70.00 & 70.00 & 82.00 & 74.00 & 90.00 & 92.00 & 70.00 & 86.00 & \textbf{75.60} \\
        VRA (+3 LVLMs) & 56.00 & 82.00 & 80.00 & 72.00 & 76.00 & 80.00 & 82.00 & 84.00 & 78.00 & 98.00 & \textbf{78.80} \\
        \bottomrule
    \end{tabular}
    \label{tab:vrsbench_accuracy}
\end{table*}

\begin{table*}[h]
    \centering
    \scriptsize
    \setlength{\tabcolsep}{4pt} 
    \caption{VRSBench VQA Subset (50 questions per type) Runtime in minutes}
    \begin{tabular}{lccccccccccc}
        \toprule
        Model & obj\_quantity & obj\_position & obj\_direction & obj\_size & reasoning & obj\_color & obj\_existence & obj\_category & obj\_shape & scene\_type & \textbf{overall} \\
        \midrule
        GeoChat & 0.79 & 0.83 & 0.79 & 0.81 & 1.05 & 0.84 & 0.81 & 0.91 & 0.78 & 1.24 & \textbf{0.89} \\
        VRA (GeoChat) & 135.26 & 145.62 & 144.87 & 133.42 & 142.09 & 144.86 & 143.71 & 141.49 & 142.27 & 134.08 & \textbf{140.77} \\
        LLaVA-1.5 & 0.56 & 0.71 & 0.84 & 0.80 & 1.08 & 0.68 & 0.74 & 0.79 & 0.73 & 0.93 & \textbf{0.79} \\
        VRA (LLaVA-1.5) & 149.32 & 155.86 & 155.37 & 157.06 & 142.53 & 139.65 & 143.76 & 145.92 & 138.23 & 143.66 & \textbf{147.14} \\
        Gemma 3 & 2.29 & 2.63 & 2.64 & 4.46 & 3.28 & 1.72 & 2.40 & 3.08 & 2.46 & 3.92 & \textbf{2.89} \\
        VRA (Gemma 3) & 170.06 & 163.17 & 183.04 & 189.95 & 185.91 & 194.74 & 174.01 & 179.46 & 179.34 & 186.15 & \textbf{180.58} \\
        VRA (+2 LVLMs) & 156.53 & 166.21 & 155.29 & 160.79 & 152.47 & 159.94 & 154.45 & 160.59 & 140.46 & 149.14 & \textbf{155.59} \\
        VRA (+3 LVLMs) & 190.03 & 196.92 & 195.75 & 212.28 & 197.24 & 184.62 & 174.63 & 197.11 & 164.74 & 177.57 & \textbf{189.09} \\
        \bottomrule
    \end{tabular}
    \label{tab:vrsbench_runtime}
\end{table*}


\subsection{Prompt Templates}
\label{sec:vra_prompt_template}

\begin{table}[h]
    \centering
    \caption{Prompt template for VRA -- LRM performs Think-Critique-Act process in Cycle 1.}
    \begin{tcolorbox}[colframe=black, colback=white, arc=8pt, boxrule=0.8pt]
    \footnotesize
    \begin{tabular}{p{0.9\textwidth}}
        {\normalsize \textbf{\textcolor{red}{System Prompt}}} \\[3pt]
        You are expert in remote sensing and geospatial image analysis. \\[6pt]
        \textbf{Task:} \\
        1. Provide a $\sim$50 word answer to the user's question based on the conversation. \\
        2. Reflect and critique your answer. \\
        3. Provide one question to ask vision model for retrieving more visual information. Your question should be straightforward and relevant to the answer and user question. \\[16pt]
        \hline
        \vspace{3pt}
        {\normalsize \textbf{\textcolor{red}{User Prompt}}} \\[3pt]
        Reflect on the user's original question and the actions taken thus far. 
    \end{tabular}
    \end{tcolorbox}
\end{table}

\begin{table}[h]
    \centering
    \caption{Prompt template for VRA -- LVLMs in the Vision-Language Suite.}
    \begin{tcolorbox}[colframe=black, colback=white, arc=8pt, boxrule=0.8pt]
    \footnotesize
    \begin{tabular}{p{0.9\textwidth}}
        {\normalsize \textbf{\textcolor{red}{System Prompt}}} \\[3pt]
        \textit{(No system prompt)} \\[16pt]
        \hline
        \vspace{3pt}
        {\normalsize \textbf{\textcolor{red}{User Prompt}}} \\[3pt]
        \textcolor{blue}{\textless image\textgreater} \textcolor{blue}{\{LRM\_generated\_question\}}
    \end{tabular}
    \end{tcolorbox}
\end{table}

\begin{table}[h]
    \centering
    \caption{Prompt template for for VRA -- LRM performs Think-Critique-Act process in Cycle $\geq 2$.}
    \begin{tcolorbox}[colframe=black, colback=white, arc=8pt, boxrule=0.8pt]
    \footnotesize
    \begin{tabular}{p{0.9\textwidth}}
        {\normalsize \textbf{\textcolor{red}{System Prompt}}} \\[3pt]
        You are expert in remote sensing and geospatial image analysis. In the preceding messages, you will find multiple tools' outputs providing visual information. \\[16pt]
        \textcolor{blue}{\{history\}} \\[6pt]
        \textbf{Task:} \\
        Revise your previous answer using the new visual information provided by multiple tools' outputs. \\
        - You should use the previous critique to add important information to your answer. \\
        \quad - You MUST include numerical citations in your revised answer to ensure it can be verified. \\
        \quad - Add a "References" section to the bottom of your answer (which does not count towards the word limit). In form of: \\
        \quad \quad - [1] visual information here \\
        \quad \quad - [2] visual information here \\
        \quad \quad - More visual information here if there is any... \\
        - You should use the previous critique to remove superfluous information from your answer and make SURE it is not more than 50 words. \\
        - You should provide one question to ask vision model for retrieving more visual information. Your question should be straightforward without repeating previous questions. \\[16pt]
        \hline
        \vspace{3pt}
        {\normalsize \textbf{\textcolor{red}{User Prompt}}} \\[3pt]
        Reflect on the user's original question and the actions taken thus far. 
    \end{tabular}
    \end{tcolorbox}
\end{table}

\begin{table}[h]
    \centering
    \caption{Prompt template for the evaluation model used in VRSBench VQA Dataset.}
    \begin{tcolorbox}[colframe=black, colback=white, arc=8pt, boxrule=0.8pt] 
    \footnotesize
    \begin{tabular}{p{0.9\textwidth}} 
        {\normalsize \textbf{\textcolor{red}{System Prompt}}} \\[3pt]
        \textit{(No system prompt)} \\[6pt]
        \hline
        \vspace{3pt}
        {\normalsize \textbf{\textcolor{red}{User Prompt}}} \\[3pt]
        Question: \textcolor{blue}{\{question\}} \\ 
        Ground Truth Answer: \textcolor{blue}{\{ground\_truth\}} \\ 
        Predicted Answer: \textcolor{blue}{\{prediction\}} \\ 
        Does the predicted answer match the ground truth? Answer 1 for match and 0 for not match. \\
        Use semantic meaning not exact match. Synonyms are also treated as a match, e.g., football and soccer, playground and ground track field, building and rooftop, pond and swimming pool. \\
        Do not explain the reason.
    \end{tabular}
    \end{tcolorbox} 
\end{table}

\end{document}